\documentclass{article} 
\usepackage{iclr2019_conference,times}
\iclrfinaltrue

\usepackage{amsmath,amsfonts,bm}

\def\eqref#1{equation~\ref{#1}}

\def\1{\bm{1}}

\DeclareMathAlphabet{\mathsfit}{\encodingdefault}{\sfdefault}{m}{sl}
\SetMathAlphabet{\mathsfit}{bold}{\encodingdefault}{\sfdefault}{bx}{n}

\usepackage{hyperref}
\usepackage{url}

\title{Deep Learning to Predict Student Outcomes}

\author{Byung-Hak Kim \\
Udacity, AI Team\\
\texttt{hak@udacity.com} \\
}

\begin{document}

\maketitle
\vspace*{-2mm}
\begin{abstract}
The increasingly fast development cycle for online course contents, along with the diverse student demographics in each online classroom, make real-time student outcomes prediction an interesting topic for both industrial research and practical needs. In this paper, we tackle the problem of real-time student performance prediction in an on-going course using a domain adaptation framework. This framework is a system trained on labeled student outcome data from previous coursework but is meant to be deployed on another course. In particular, we introduce a GritNet architecture, and develop an \emph{unsupervised} domain adaptation method to transfer a GritNet trained on a past course to a new course without any student outcome label. Our results for real Udacity student graduation predictions show that the GritNet not only \emph{generalizes} well from one course to another across different Nanodegree programs, but also enhances real-time predictions explicitly in the first few weeks when accurate predictions are most challenging. 
\end{abstract}
\vspace*{-2mm}

\section{Introduction}
\label{sec:introduction}
\vspace*{-2mm}
With the growing need for people to keep learning throughout their careers, massive open online course (MOOCs) companies, such as Udacity and Coursera, not only aggressively design new and relevant courses, but they also refresh existing course content to keep the material up-to-date. This effort results in a significant increase in student numbers, which make it impractical for even experienced human instructors to assess an individual student and anticipate their learning outcomes. Moreover, students in each MOOC classroom are heterogeneous in their background and intention, which is very different from a classic classroom \citep{Chuang16, Economist17}. Even subsequent offerings of a course within a year can have different sets of students, mentors, instructors, content, and workflows. In this world of MOOCs, an automated machine which reliably forecasts a student's performance early in their coursework would be a valuable tool for making smart decisions about when (and with whom) to make live educational interventions. These interventions have the aim of increasing engagement, providing motivation, and empowering students to succeed. 
\vspace*{-2mm}

\section{GritNet}
\label{sec:GritNet}
\vspace*{-2mm}
In that, we first recast the student performance prediction problem as a sequential event prediction problem and introduce the GritNet. The task of predicting student outcomes can be expressed as a sequential event prediction problem: given a past event sequence \(\mathbf{o}\triangleq(o_{1},\dots,o_{T})\) taken by a student, estimate likelihood of future event sequence \(\mathbf{y}\triangleq(y_{T+D},\dots,y_{T'})\) where \(D \in \mathbb Z_{+}\). In the form of online classes, each event \(o_{t}\) represents a student's action (or activities) associated with a time stamp. In other words, \(o_{t}\) is defined as a paired tuple of \((a_{t},d_{t})\), where each action \(a_{t}\) represents, for example, ``a lecture video viewed'', ``a quiz answered correctly/incorrectly'', or ``a project submitted and passed/failed'', and \(d_{t}\) states the corresponding (logged) time stamp. Then, the log-likelihood of \(p(\mathbf{y}|\mathbf{o})\) can be written approximately as \(\sum_{i=T+D}^{T'} \log p(y_{i}|\upsilon)\), given a fixed-dimensional embedding representation \(\upsilon\) of \(\mathbf{o}\). The goal of each GritNet is, therefore, to compute an individual log-likelihood \(\log p(y_{i}|\upsilon)\), and those estimated scores can be simply added up to estimate long-term student outcomes.

In order to feed students' raw event records into the GritNet, it is necessary to encode the time-stamped logs (ordered sequentially) into a sequence of fixed-length input vectors by \emph{one-hot encoding}. A one-hot vector is a vector \(\mathbbm{1}(a_{t}) \in \{0,1\}^L\), where \(L\) is the number of unique actions and only one element could take the value 1 to distinguish each activity \(a_{t}\) from every other. One-hot vectors of the same student are connected into a long vector sequence to represent the student's whole sequential activities in \(\mathbf{o}\). To further capture a student's varying learning speeds, GritNet defines the discretized time difference between adjacent events \(\Delta_{t}\) as \(d_{t}-d_{t-1}\)\footnote{For the Udacity data described in Section~\ref{ssec:Udacity Data}, we use day to represent inter-event time intervals.}. Then, one-hot encode \(\Delta_{t}\) into \(\mathbbm{1}(\Delta_{t})\) and connect them with the corresponding \(\mathbbm{1}(a_{t})\) to represent \(\mathbbm{1}(o_{t})\) as \([\mathbbm{1}(a_{t});\mathbbm{1}(\Delta_{t})]\). Lastly, the output sequences shorter than the maximum event sequence length (of a given training set) are pre-padded with all \(\mathbf{0}\) vectors.

The complete GritNet architecture is the following. The first embedding layer \citep{Bengio01} learns an embedding matrix \(\mathbf{E}^o \in \mathbb{R}^{E\times|O|}\), where \(E\) and \(|O|\) are the embedding dimension and the number of unique events (i.e., input vector \(\mathbbm{1}(o_{t})\) size). This embedding can be used to convert an input vector \(\mathbbm{1}(o_{t})\) onto a low-dimensional embedding \(\mathbf{\upsilon}_{t}\) defined as \(\mathbf{E}^o\mathbbm{1}(o_{t})\). This (dense) event embedding \(\mathbf{\upsilon}_{t}\) is then passed into the bidirectional long short term memory (BLSTM) \citep{Graves05} and the output vectors are formed by concatenating each forward and backward direction outputs. Next, a global max pooling (GMP) layer \citep{Collobert08} is added to form a fixed dimension vector (independent of the input event sequence length) by taking the maximum over time (over the input event sequence). This GMP layer is able to capture the most relevant signals over the sequence and is also able to deal with the imbalanced nature of data. One can view this GMP layer as a hard self-attention layer or can consider the GMP layer output generates a sequence level embedding of the whole input event sequence. The GMP layer output is, ultimately, fed into a fully-connected (FC) layer, and a softmax (i.e., sigmoid) layer subsequently, to calculate the log-likelihood \(\log p(y_{i}|\upsilon)\).

In particular, GritNet is the first deep learning architecture which successfully advances the state of the art by demonstrating substantial prediction accuracy improvements (particularly pronounced in the first few weeks when predictions are extremely challenging). In contrast to other works \citep{Mi15, Piech15, Whitehill17, Wang17}, the GritNet does not need any feature engineering (it can learn from raw input) and it can operate on any (raw) student event data associated with a time stamp even when highly imbalanced.
\vspace*{-2mm}

\section{Domain Adaptation with GritNet}
\label{sec:Domain Adaptation with GritNet}
\vspace*{-2mm}
Though GritNet's superior prediction accuracy was reported in \citet{Kim18}, it was not fully addressed whether GritNet models transfer well to different courses or if they could be deployed for real-time prediction with ongoing courses. With an increase in new courses and the fast pace of content revisions of MOOC classes to meet students' educational needs, the GritNet model is required to generalize well to unseen courses and to predict performance in real-time. This prediction happens for students who have not yet finished the course (i.e., new data without prior knowledge of the students' outcomes). 
\vspace*{-2mm}

\subsection{Ordinal Input Encoding}
\label{ssec:Ordinal Input Encoding}
\vspace*{-2mm}
Specifically, in order to leverage the unlabeled data in the target course, we propose an ordinal input encoding procedure as a basis for GritNet to provide transferable sequence-level embeddings across courses. For any MOOC coursework, each content exhibits a natural order, following a contents tree. For the sake of simplicity, we group content IDs into subcategories and convert each content ID to corresponding ordinal ID by exploiting actual ordering implicit in the content paths within each category. Then each encoded MOOC contains the same number of unique actions as the original. For example, the Udacity data in Table~\ref{table:Udacity data characteristic} contains three subcategories - content pages, quizzes, and projects. One Nanodegree program dataset has three sequences of ordinal IDs - content-\(1\) to content-\(i\), quiz-\(1\) to quiz-\(j\) and project-\(1\) to project-\(k\). Note that with this encoding, the number of unique actions \(L\) turns out to be equivalent to \(i+2\times j+2\times k\), since quizzes and projects allow two potential actions (i.e. a quiz answered correctly/incorrectly or a project submitted and passed/failed). It is also worth mentioning that different courses may have different $i$, $j$, $k$ and $L$ values.
\vspace*{-2mm}

\subsection{Pseudo-Labels and Transfer Learning}
\label{ssec:Pseudo Labels and Transfer Learning}
\vspace*{-2mm}
To deal with unlabeled target course data in \emph{unsupervised} domain adaptation \citep{Pan10} from source to target course, we harness the trained GritNet on a source data to assign pseudo labels to the target data. This is a common real-world scenario when releasing new, never-before consumed educational content. In other words, we use a trained GritNet on a source course to assign pseudo outcome labels to the target course, then continue training the GritNet on the target course (as precisely described in Algorithm \ref{alg:algorithm1}). 

\vspace*{-2mm}
\algnewcommand\algorithmicreturn{\textbf{return}}
\algnewcommand\RETURN{\algorithmicreturn}
\algnewcommand\algorithmicprocedure{\textbf{procedure}}
\algnewcommand\PROCEDURE{\item[\algorithmicprocedure]}%
\algnewcommand\algorithmicendprocedure{\textbf{end procedure}}
\algnewcommand\ENDPROCEDURE{\item[\algorithmicendprocedure]}%
\algnewcommand{\algvar}[1]{{\text{\ttfamily\detokenize{#1}}}}
\algnewcommand{\algarg}[1]{{\text{\ttfamily\itshape\detokenize{#1}}}}
\algnewcommand{\algproc}[1]{{\text{\ttfamily\detokenize{#1}}}}
\algnewcommand{\algassign}{\leftarrow}

\begin{algorithm*}[ht]
    \caption{Domain Adaptation with GritNet}
    \label{alg:algorithm1}
    \begin{algorithmic}[1]
        \raggedright
        \REQUIRE Source course data \(\mathcal{X}_{source}\), Source label \(\mathcal{Y}_{source}\), Target data \(\mathcal{X}_{target}\), Threshold \(\theta\)
        \STATE Set source training set as \(\mathcal{T}_{source}=(\mathcal{X}_{source}, \mathcal{Y}_{source})\) \label{step1}
        \STATE Train \({\textit{GritNet}}_{source}\) with \(\mathcal{T}_{source}\) \label{step2}
        \STATE Evaluate on \(\mathcal{X}_{target}\): \(\mathcal{\hat{Y}}_{pred}\):=\({\textit{GritNet}}_{source}\)(\(\mathcal{X}_{target}\)) \label{step3}
        \STATE Assign pseudo-labels to \(\mathcal{X}_{target}\): \(\mathcal{Y}_{label}\):=\({\mathbbm{1}}\)(\(\mathcal{\hat{Y}}_{pred}\geq\theta\)) \label{step4}
        \STATE Update target training set as \(\mathcal{T}_{adapt}=(\mathcal{X}_{target}, \mathcal{Y}_{label})\) \label{step5}
        \STATE Freeze all the \({\textit{GritNet}}_{source}\) but the last FC layer and continue training with \(\mathcal{T}_{adapt}\)\label{step6}
    \end{algorithmic}
\end{algorithm*}

\vspace*{-2mm}
This method was motivated by the observation that the GMP layer output captures generalizable sequential representation of an input event sequence, whereas the high-level FC layer learns the course-specific features. It is worth mentioning that the last FC layer limits the number of parameters to learn in Step~\ref{step6}. So this enables the entire Algorithm \ref{alg:algorithm1} to be a very practical solution for faster adaptation time (and in fact deployment cycle) and applicability for even a (relatively) smaller size target course. 
To our knowledge, this is the first work to effectively answer the source and target course distribution mismatch encountered in real-time student performance prediction, which had been considered as a challenging open problem in prior works \citep{Boyer15, Whitehill17, Whitehill17b, Dalipi18}.
\vspace*{-2mm}

\section{Generalization Performance}
\label{sec:Generalization Performance}
\vspace*{-2mm}
\subsection{Udacity Data}
\label{ssec:Udacity Data}
\vspace*{-2mm}
To demonstrate the benefits of Algorithm~\ref{alg:algorithm1}, we benchmark the student graduation prediction task on various Udacity dataset (see Table~\ref{table:Udacity data characteristic} for more information of each Udacity dataset used for this study). In all programs, graduation is defined as completing each of the required projects in a program's curriculum with a passing grade. Each program has a designated calendar start and end date for each cohort of students that passes through, and users have to graduate before the official end of their cohort's term to be considered successfully graduated. Each program's curriculum contains a mixture of video contents, written contents, quizzes, and projects. Note that it is not required to interact with every piece of content or complete every quiz to graduate. For all datasets, an event represents a user taking a specific action (e.g., watching a video, reading a text page, attempting a quiz, or receiving a grade on a project) at a certain time stamp. It should be noted that no personally identifiable information is included in this data and student equality is determined via opaque unique ids.

\newcolumntype{C}{>{\raggedleft\arraybackslash}X} 
\begin{table*}
  \caption{Udacity data characteristics}
  \vspace*{-2mm}
  \label{table:Udacity data characteristic}
  \centering
  \begin{tabularx}{\textwidth}{@{}l*{11}{C}c@{}}
    Dataset & Enrolled From & Students & Contents (i) & Quizzes (j) & Projects (k) & AvgSeq. Length & Graduates & Grad. Rates \\ 
    \midrule
    ND-A v1 & 2/3/2015 & 5,626   & 471 & 168 & 4 & 421 & 1,202 & 21.4\% \\ 
    ND-A v2 & 4/1/2015 & 2,230   & 471 & 168 & 4 & 881 & 453 & 20.3\% \\ 
    ND-B & 3/22/2017 & 4,377   & 568 & 84 & 10 & 675 & 1,726 & 39.4\% \\ 
    ND-C & 1/14/2017 & 4,761   & 346 & 50 & 5 & 430 & 2,198 & 46.2\%  \\ 
  \end{tabularx}
\end{table*}

\vspace*{-\baselineskip}
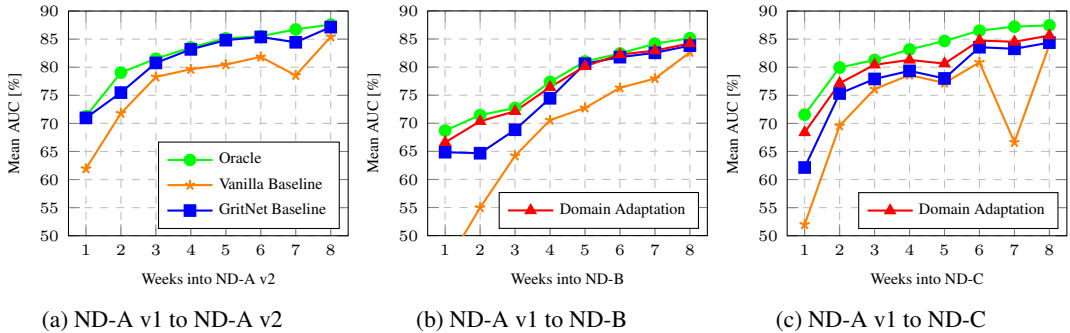
\begin{figure*}[ht]
\begin{subfigure}[ht]{0.31\textwidth}
\begin{tikzpicture}

\begin{axis}[
    xlabel={Weeks into ND-A v2}, 
    ylabel={Mean AUC [\%]},
    xmin=0.5, xmax=8.5, xtick={1,2,...,7,8},
    ymin=50, ymax=90, ytick={50,55,...,85,90},    
    xmajorgrids=true, 
    ymajorgrids=true, 
    grid style=dashed,
    label style={font=\tiny}, 
    tick label style={font=\tiny},  
    legend pos=south east,
    legend style={font=\tiny},
    legend cell align={left}
]

\addplot[color=green,mark=*,line width=0.8pt]
    coordinates {(1,71.28)
                (2,79.04)
                (3,81.51)
                (4,83.52)
                (5,85.13)
                (6,85.52)
                (7,86.73)
                (8,87.59)};

\addplot[color=orange,mark=star,line width=0.8pt]
    coordinates {(1,61.94)
                (2,71.79)
                (3,78.27)
                (4,79.64)
                (5,80.46)
                (6,81.84)
                (7,78.54)
                (8,85.42)};

\addplot[color=blue,mark=square*,line width=0.8pt]
    coordinates {(1,70.98)
                (2,75.48)
                (3,80.77)
                (4,83.18)
                (5,84.82)
                (6,85.38)
                (7,84.44)
                (8,87.14)};

\legend{Oracle, Vanilla Baseline, GritNet Baseline}
\end{axis}

\end{tikzpicture}
\vspace*{-\baselineskip}
\caption{ND-A v1 to ND-A v2} \label{fig:performance-1a}
\end{subfigure}
\hspace*{\fill} 
\begin{subfigure}[ht]{0.31\textwidth}
\begin{tikzpicture}

\begin{axis}[
    xlabel={Weeks into ND-B}, 
    ylabel={Mean AUC [\%]},
    xmin=0.5, xmax=8.5, xtick={1,2,...,7,8},
    ymin=50, ymax=90, ytick={50,55,...,95,100},
    xmajorgrids=true, 
    ymajorgrids=true, 
    grid style=dashed,
    label style={font=\tiny}, 
    tick label style={font=\tiny},  
    legend pos=south east,
    legend style={font=\tiny},
    legend cell align={left}
]

\addplot[color=green,mark=*,line width=0.8pt]
    coordinates {(1,68.69)
                (2,71.45)
                (3,72.72)
                (4,77.36)
                (5,80.99)
                (6,82.42)
                (7,84.16)
                (8,85.15)};

\addplot[color=orange,mark=star,line width=0.8pt]
    coordinates {(1,44.54)
                (2,54.98)
                (3,64.21)
                (4,70.56)
                (5,72.73)
                (6,76.31)
                (7,77.96)
                (8,82.63)};

\addplot[color=blue,mark=square*,line width=0.8pt]
    coordinates {(1,64.85)
                (2,64.67)
                (3,68.85)
                (4,74.45)
                (5,80.60)
                (6,81.77)
                (7,82.54)
                (8,83.82)};

\addplot[color=red,mark=triangle*,line width=0.8pt]
    coordinates {(1,66.60)
                (2,70.36)
                (3,72.14)
                (4,76.42)
                (5,80.15)
                (6,82.24)
                (7,82.93)
                (8,84.22)};

\legend{,,,Domain Adaptation}
\end{axis}

\end{tikzpicture}
\vspace*{-\baselineskip}
\caption{ND-A v1 to ND-B} \label{fig:performance-1b}
\end{subfigure}
\hspace*{\fill} 
\begin{subfigure}[ht]{0.31\textwidth}
\begin{tikzpicture}

\begin{axis}[
    xlabel={Weeks into ND-C}, 
    ylabel={Mean AUC [\%]},
    xmin=0.5, xmax=8.5, xtick={1,2,...,7,8},
    ymin=50, ymax=90, ytick={50,55,...,95,100},
    xmajorgrids=true, 
    ymajorgrids=true, 
    grid style=dashed,
    label style={font=\tiny}, 
    tick label style={font=\tiny},  
    legend pos=south east,
    legend style={font=\tiny},
    legend cell align={left}
]

\addplot[color=green,mark=*,line width=0.8pt]
    coordinates {(1,71.56)
                (2,79.95)
                (3,81.28)
                (4,83.16)
                (5,84.67)
                (6,86.51)
                (7,87.22)
                (8,87.46)};

\addplot[color=orange,mark=star,line width=0.8pt]
    coordinates {(1,52.00)
                (2,69.57)
                (3,76.11)
                (4,78.61)
                (5,77.19)
                (6,80.86)
                (7,66.63)
                (8,83.99)};

\addplot[color=blue,mark=square*,line width=0.8pt]
    coordinates {(1,62.15)
                (2,75.32)
                (3,77.91)
                (4,79.35)
                (5,78.00)
                (6,83.56)
                (7,83.28)
                (8,84.38)};

\addplot[color=red,mark=triangle*,line width=0.8pt]
    coordinates {(1,68.39)
                (2,77.15)
                (3,80.44)
                (4,81.30)
                (5,80.64)
                (6,84.76)
                (7,84.52)
                (8,85.64)};

\legend{,,,Domain Adaptation}

\end{axis}

\end{tikzpicture}
\vspace*{-\baselineskip}
\caption{ND-A v1 to ND-C} \label{fig:performance-1c}
\end{subfigure}
\vspace*{-2mm}
\caption{Real-time student graduation prediction accuracy comparisons of models (\subref{fig:performance-1a}) from ND-A v1 to the next version ND-A v2 and (\subref{fig:performance-1b}-\subref{fig:performance-1c}) from earlier ND-A v1 program to two later Udacity Nanodegree programs, ND-B and ND-C. (\subref{fig:performance-1a}): GritNet baseline shows only 1.01\% abs accuracy loss ($<$ 5.30\% abs loss of vanilla baseline) in average over eight weeks as compared to oracle bound. The accuracy gains of GritNet baseline suggest that sequence-level embeddings trained with GritNet result in more transferable features as compared to features learned with a Vanilla baseline model. (\subref{fig:performance-1b}-\subref{fig:performance-1c}): Being consistent with (\subref{fig:performance-1a}) results, the clear wins of the GritNet baseline over the Vanilla baseline reaffirm that the sequence-level embedding trained with GritNet is more robust to source and target course distribution mismatch as compared to features learned with the Vanilla baseline model. Furthermore, domain adaptation provides 70.60\% accuracy recovery in average during first four weeks (up to 84.88\% at week 3) for ND-B dataset and 58.06\% accuracy recovery during the same four weeks (up to 75.07\% at week 3) for ND-C dataset from GritNet baseline performances.}
\label{fig:performance-1}
\end{figure*}

\subsection{Performance Results}
\label{ssec:Performance Results}
\vspace*{-2mm}
In order to evaluate the transferability of the GritNet across different courses, we consider three different scenarios from one course to another on the setups below (see Figure~\ref{fig:performance-1} for the results):
\vspace*{-2mm}
\begin{itemize}
    \item \textbf{Vanilla Baseline} using the same logistic regression based baseline model in \citet{Kim18} is trained on a source course and evaluated on a target one. \vspace*{-1.25mm}
    \item \textbf{GritNet Baseline} is trained on a source course and evaluated on a target course. Note that there is no training data or labels from target course for training the GritNet. \vspace*{-1.25mm}
    \item \textbf{Domain Adaptation} follows the steps specified in Algorithm~\ref{alg:algorithm1}. Note that for pseudo-labels generation (Step~\ref{step4}), a hyper-parameter \(\theta\) threshold of \([0.1, 0.2, 0.3,0.4]\) is tried. \vspace*{-1.25mm}
    \item \textbf{Oracle} performance bound of domain adaption is computed as the same procedures in Algorithm~\ref{alg:algorithm1} while skipping Step~\ref{step3} and Step~\ref{step4} to use oracle target labels \(\mathcal{Y}_{target}\) instead of assigned pseudo-labels \(\mathcal{Y}_{label}\). \vspace*{-1.25mm}
\end{itemize}
\vspace*{-2mm}
For all models across the setups, we trained different models of a GritNet architecture of BLSTM with forward and backwards LSTM layers containing 256 cell dimensions per direction and embedding layer dimension of 512 for different weeks. Each model is based on students' week-by-week event records, to predict whether each student was likely to graduate. Further hyper-parameter optimization could be done for the optimal accuracy at each week.\vspace*{-1mm}

The accuracy of each system's prediction was measured by the area under the Receiver Operating Characteristic curve (AUC) which scores between 0 and 100\% (the higher, the better)\footnote{Since the true binary target label (1: graduate, 0: not graduate) is imbalanced (i.e., number of 0s outweighs number of 1s), accuracy is not an appropriate metric.}. We used \(5\)-fold student level cross-validation, while ensuring each fold contained roughly the same proportions of the two groups (graduate and non graduate) of students. To compare domain adaption performance with the upper bound assessed with oracle setup, we consider mean AUC recovery rate (ARR), as a metric to evaluate AUC improvements from \emph{unsupervised} domain adaptation against oracle training. This measure is defined as absolute AUC reduction that \emph{unsupervised} domain adaptation produces divided by the absolute reduction that oracle (\emph{supervised}) training produces. We assume that the upper bound for \emph{unsupervised} domain adaptation would be the performance of the oracle training on the same data, which has a AUC recovery of 100\%. \vspace*{-2mm}

\section{Conclusion}
\label{sec:Conclusion}
\vspace*{-2mm}
In this paper, we have proposed an extremely practical approach for predicting real-time student performance, which has been considered as of the utmost importance in MOOCs (but under-explored), as this enables student outcomes prediction while a course is on-going. We introduced a novel domain adaptation algorithm with GritNet and demonstrated that GritNet can be transferred to new courses without labels and provides a substantial AUC recovery rate. This method is effective in the sense that it works across different courses varying in lengths, format and contents and does not require custom feature engineering or additional target-course data or labels. Encouraged by this result, many future directions are feasible to explore. One potential direction is to look into a GritNet pretraining across more diverse courses. Given a pretrained GritNet and fine-tuning on the target course, it would boost performance further. We hope that our results will catalyze new developments of transfer learning for the real-time student performance prediction problem.

\bibliography{iclr2019_conference}
\bibliographystyle{iclr2019_conference}

\end{document}